\documentclass[10pt,twocolumn,letterpaper]{article}

\usepackage{cvpr}              
\usepackage{makecell}

\usepackage[dvipsnames]{xcolor}

\definecolor{cvprblue}{rgb}{0.21,0.49,0.74}
\usepackage[pagebackref,breaklinks,colorlinks,citecolor=cvprblue]{hyperref}

\def\paperID{64} 
\def\confName{EarthVision}
\def\confYear{2024}

\title{
Classifying geospatial objects from multiview aerial imagery using semantic meshes}

\author{David Russell\\
University of California, Davis\\
{\tt\small djrussell@ucdavis.edu}
\and
Ben Weinstein\\
University of Florida\\
{\tt\small
bweinstein@ufl.edu
}
\and
David Wettergreen\\
Carnegie Mellon University\\
{\tt\small dsw@cmu.edu}
\and
Derek Young\\
University of California, Davis\\
{\tt\small djyoung@ucdavis.edu}
}

\begin{document}
\maketitle
\begin{abstract}
Aerial imagery is increasingly used in Earth science and natural resource management as a complement to labor-intensive ground-based surveys. Aerial systems can collect overlapping images that provide multiple views of each location from different perspectives. However, most prediction approaches (\eg for tree species classification) use a single, synthesized top-down ``orthomosaic'' image as input that contains little to no information about the vertical aspects of objects and may include processing artifacts. We propose an alternate approach that generates predictions directly on the raw images and accurately maps these predictions into geospatial coordinates using semantic meshes. This method---released as a user-friendly open-source toolkit\footnote{\url{https://github.com/open-forest-observatory/geograypher}} \cite{david_russell_2024_11193027}---enables analysts to use the highest quality data for predictions, capture information about the sides of objects, and leverage multiple viewpoints of each location for added robustness. 
We demonstrate the value of this approach on a new benchmark dataset\footnote{\url{https://osf.io/6snfq/}} \cite{Russell_data_OSF_2024} of four forest sites in the western U.S. that consists of drone images, photogrammetry results, predicted tree locations, and species classification data derived from manual surveys. We show that our proposed multiview method improves classification accuracy from 53\% to 75\% relative to an orthomosaic baseline on a challenging cross-site tree species classification task.
\end{abstract}    
\section{Introduction}
\label{sec:intro}
\begin{figure}
   \begin{subfigure}{0.34\linewidth}
   \includegraphics[height=82pt]{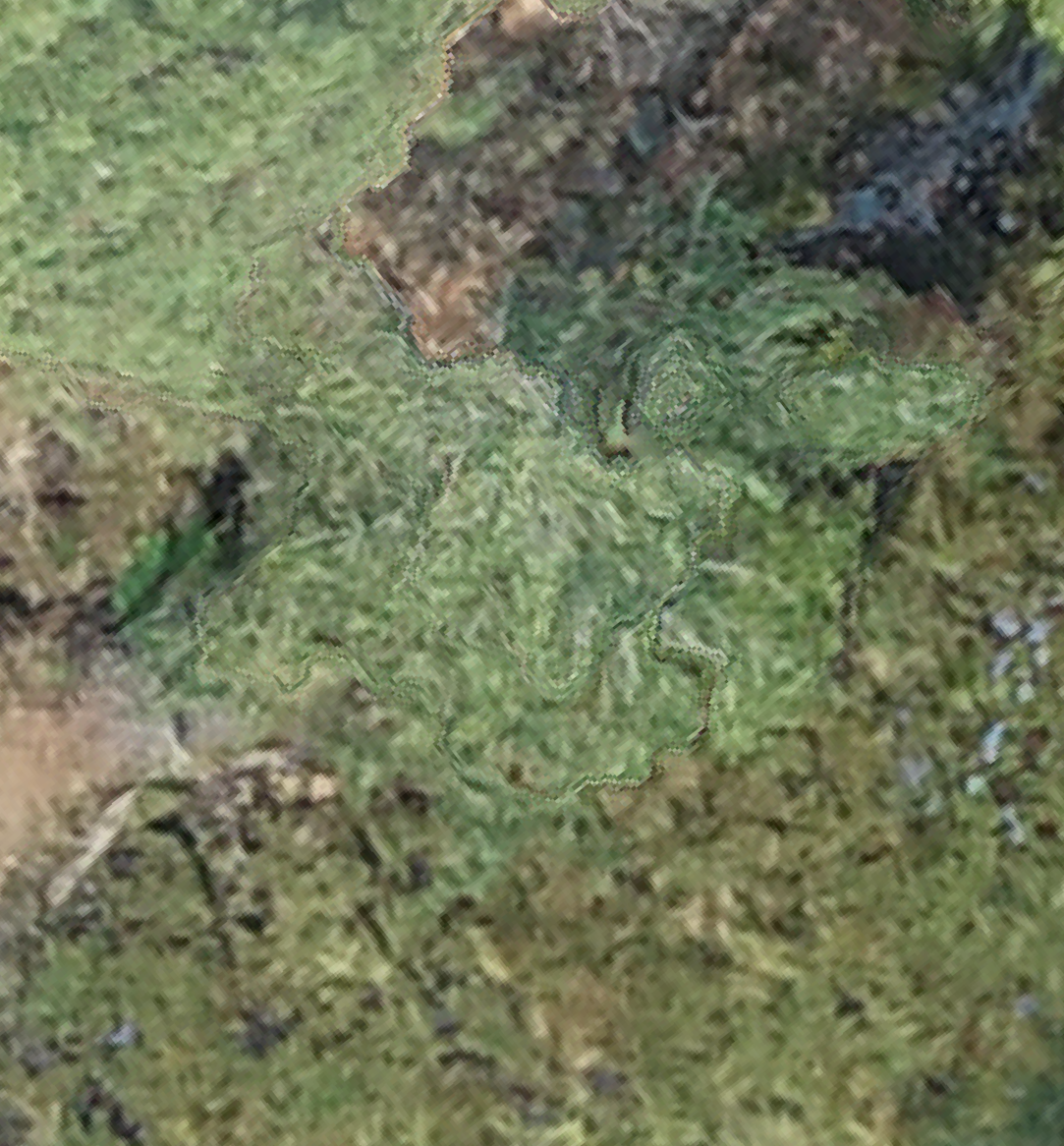}
   \caption{Orthomosaic}
   \label{fig:ortho_artifacts:ortho}
   \end{subfigure}
   \begin{subfigure}{0.32\linewidth}
   \includegraphics[height=82pt]{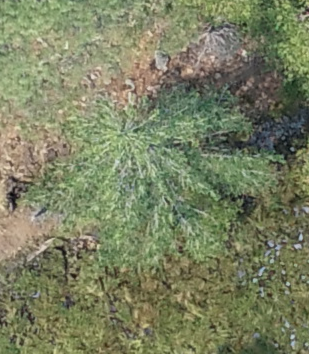}
   \caption{Raw nadir}
   \label{fig:ortho_artifacts:nadir}
   \end{subfigure}
   \begin{subfigure}{0.28\linewidth}
   \includegraphics[height=82pt]{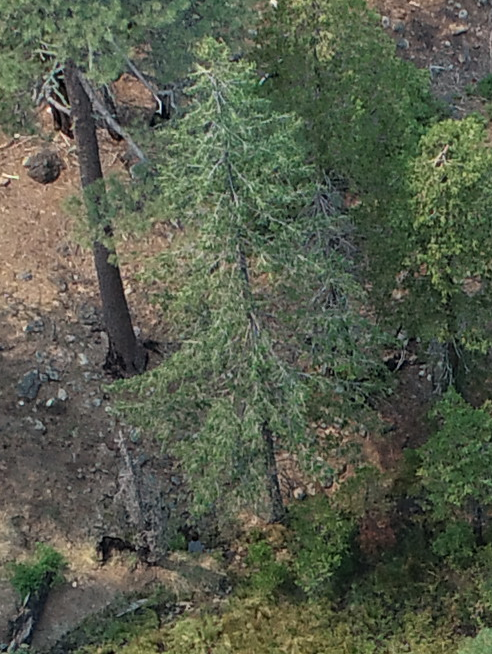}
   \caption{
   Raw oblique
   }
   \label{fig:ortho_artifacts:oblique}
   \end{subfigure}
  \caption{
  The same tree in (a) an orthomosaic created from nadir drone images collected from 120 m above ground level, (b) a raw nadir drone image collected from 120 m above ground level, and (c) a raw oblique drone image collected from 80 m above ground level. Distortions due to orthorectification are apparent in (a).
  }
  \label{fig:ortho_artifacts}
\end{figure}

Addressing questions in Earth science and natural resource management often requires detailed knowledge of the locations and identities of resources---for example, trees, birds, nest sites, and geologic structures. Traditionally, this information is collected through time- and labor-intensive ground-based manual surveys, which have limited spatial extent. To overcome these constraints, modern natural resource inventory approaches have increasingly leveraged aerial and satellite imagery to derive broad-extent, fine-scale maps of the resources of interest \cite{https://doi.org/10.3334/ornldaac/1832, Hodgson2016, SWAYZE2021112540, https://doi.org/10.1111/mms.12586}.

Aerial platforms can quickly collect many overlapping images of a region and are valuable because they can be flown on demand and have high spatial resolution.
The traditional approach to analyzing the data from these multiview surveys is to first generate an ``orthomosaic'', representing a synthesized top-down (orthographic) perspective of the scene. Orthomosaic-based analyses have enabled mapping efforts including individual tree detection \cite{Weinstein2020}, classification \cite{rs10081218}, and trait prediction \cite{Marconi21traits}. Orthomosaic-based approaches, however, have two important shortcomings. First, the orthorectification process necessarily introduces some distortion of the scene (\cref{fig:ortho_artifacts:ortho}), since the raw 2D images must be warped to the orthographic perspective and stitched together. Second, the orthomosaic only contains a top-down view, thus foregoing any opportunity to leverage multiple (e.g., oblique) perspectives that likely contain substantial information to support object detection and characterization (\cref{fig:ortho_artifacts:nadir}, \cref{fig:ortho_artifacts:oblique}).


Using the information contained in aerial images with multiple views of each object can substantially improve classification accuracy over an orthomosaic-based approach \cite{LIU2018122, Koukal2012multi_angle}. This benefit has been demonstrated most clearly in the context of wetland detection and segmentation, in which researchers achieved classification accuracy of 82\%, relative to 65\% for a baseline orthoimagery approach \cite{LIU2018122}. While useful for wetland mapping and agricultural applications \cite{rs13132622}, current approaches use a 2.5D representation of the scene and require the 3D environment to be relatively flat and free of structures that may create occlusions. This assumption does not hold in systems with substantial surface relief, such as many forests. 

To take advantage of multiview imaging while accommodating complex surface relief, we adopt a 3D mesh-based approach to geometric reasoning that allows us to accurately account for inter-object occlusions. Using this mesh representation, we aggregate segmentation predictions from a deep learning model from each viewpoint to generate a classified mesh, and we then use this mesh to classify individual trees to species. The contributions of this work are as follows:
\begin{itemize}
    \item A novel workflow for geospatial object classification using semantic meshes and multiview imagery that yields 75\% accuracy in a tree species classification experiment, compared to 53\% for an orthomosaic baseline
    \item A field-validated dataset to promote research on multiview tree species classification 
    \item A user-friendly open-source toolkit for generating landscape-scale geospatial predictions with multiview imagery
\end{itemize}


\section{Related Work}
\label{sec:related}
\subsection{Multiview reasoning in natural environments}
\label{sec:related:multi_view}
Initial efforts to apply multiview reasoning to Earth observation data focused on detailed characterization of spectral reflectance by measuring light reflected at multiple angles from a given surface (Koukal \etal \cite{Koukal2012multi_angle}, Liu \etal \cite{LIU2018122}). More recently, focus has shifted to incorporating multi-perspective imagery into computer vision-based semantic segmentation and classification routines that account for spatial context around the pixel or object of interest. Such methods have enabled improved segmentation and reconstruction of urban scenes using multiview orthorectified multispectral satellite imagery \cite{Purri2019, Leotta2019}. One approach developed for wetlands involved projecting the bounds of geospatial objects (unclassified land cover units) onto the raw drone images, cropping out each raw view of each object as its own image, classifying the resulting images using a computer vision algorithm, and aggregating the multiple predictions (one per drone image the object appeared in) to a single prediction per object using vote counting \cite{rs10030457}. 
An open-source tool called EasyIDP\footnote{\url{https://github.com/UTokyo-FieldPhenomics-Lab/EasyIDP}} \cite{rs13132622} aims to facilitate a similar workflow for agricultural applications by projecting the bounds of a given geospatial region onto the aerial images. However, these approaches do not translate well to high-relief systems (e.g. forest canopies) and/or oblique images because (a) they represent surface relief using a 2.5D digital elevation model that cannot capture undercut regions, and (b) they do not account for the possibility of occluding structures.

The works of Russell \etal \cite{Russell2022Mapping} and Andrada \etal \cite{Andrada2023Semantic} expand beyond photogrammetry-based methods by using a drone equipped with a lidar sensor, an inertial measurement unit, and cameras to classify and map forest fuels in a structurally-complex scene. These approaches use deep learning applied to the camera images to classify fuel and use the lidar to identify its 3D location relative to the drone platform. To build a globally-coherent map from multiple observations, these approaches use simultaneous localization and mapping (SLAM) \cite{SLAM} to build a real-time estimate of the drone's location and thereby the structure of the world. Unfortunately, SLAM methods are susceptible to both accumulating errors and catastrophic failures in localization. This can result in maps that while locally consistent, do not accurately represent the true geospatial location of objects, and they are therefore challenging to use for natural resource management and analysis tasks. 

One option for representing complex geometries is the mesh data structure that is commonly used in computer graphics \cite{Buss2009}. Such meshes are generally produced using structure-from-motion photogrammetry, which uses a set of overlapping images to predict (a) camera lens parameters, (b) the pose (position and orientation) of the camera at each image capture, and (c) a 3D representation of the scene constructed via triangulation of keypoints common across images. 
Meshes consist of \textit{vertices} (points in 3D space), connected by triangular \textit{faces}. The geometric information contained in the mesh data structure can be augmented by adding additional ``semantic'' information to the faces or vertex such as a species ID for that location.
Work such as \cite{Valentin2013} generates semantic meshes using per-image predictions by first classifying each pixel of the input images and then associating these predictions to locations on the mesh surface. This approach is validated on indoor and street-scale scenes, but to our knowledge, no prior work uses semantic meshes and un-rectified multiview image-based predictions for large-scale aerial mapping tasks.

\subsection{Predicting tree species using remote sensing imagery}
\label{sec:related:tree_species}
Classification of tree species using remote sensing data has been a highly active field of research for at least two decades \cite{FASSNACHT201664}. An initial focus on multispectral satellite imagery has expanded in recent years to include species prediction from aerial imagery, generally driven by computer vision algorithms employing artificial neural networks. For example, researchers have used fine-tuned CNNs, applied to RGB (red, green, blue) drone-derived orthoimagery, to identify the species of Amazonian palms \cite{FERREIRA2020118397} and seven Japanese tree species \cite{Onishi_tree_classification}. Rather than using orthomosaics, Santos \etal \cite{s19163595} applied CNNs to un-rectified drone imagery to detect specific tree species of concern among urban trees. However, these experiments were only conducted at the image-level and predictions were not mapped into geospatial coordinates, which is critical for many management applications. Approaches to species classification include (a) first cropping out individual (unclassified) tree crowns detected by other methods and then applying a classification algorithm \cite{Onishi_tree_classification}, (b) applying pixel-level semantic segmentation to orthoimagery and then defining objects as clusters of pixels of the same class 
\cite{FERREIRA2020118397} or (c) applying architectures such as YOLO \cite{redmon2016look} that simultaneously detect objects and classify them \cite{s19163595}. Most often, species classes are obtained through manual, direct annotation of images by experts \cite{FERREIRA2020118397, s19163595}, which may inflate accuracy because human annotators observe the same images as the vision algorithm and may make similar errors. Using a more robust approach of assigning tree species classes through field-based surveys, researchers classified 72 species in a continental-scale survey using pixel-based classification of crewed-aircraft hyperspectral imagery (426 reflectance bands) \cite{MARCONI2022113264}. Later work built on this approach to more accurately classify species by using multi-temporal data and hierarchical computer vision-based classification \cite{Weinstein2023long_tail}. Although some tree species classification work has leveraged imagery of the same trees from multiple time points and multiple sensors \cite{Weinstein2023long_tail, Ahlswede2023TreeSatAI}, no work to our knowledge has used multiple views of the same trees from different angles. 



\section{Dataset}
\label{sec:dataset}
\begin{table*}
\begin{center}
    \begin{tabular}{||c c c c c c c||} 
     \hline
     Name & Coordinates & \makecell{Fire\\year} & \makecell{Imagery\\year} & \makecell{Field data\\year} & \makecell{Reference\\tree count} & \makecell{Dominant\\species} \\ [0.5ex] 
     \hline\hline
     Delta & 41.04, -122.53 & 2018 & 2020 & 2020, 2021 & 288 & PIPJ, PSME, PILA \\ 
     \hline
     Chips & 40.13, -121.11 & 2012 & 2020, 2021 & 2020 & 243 & PSME, PIPJ, ABCO \\
     \hline
     Valley & 38.82, -122.69 & 2015 & 2020 & 2020 & 64 & PIPJ, PSME, PILA  \\
     \hline
     Lassic & 40.30, -123.57 & 2015 & 2021 & 2020, 2021 & 178 & CADE, ABCO, PIPJ \\
     \hline
    \end{tabular}
    \end{center}
    \caption{Study site attributes. Coordinates are provided as latitude, longitude. Dominant species are listed in order of decreasing abundance.}
    \label{tab:site_attributes}
\end{table*}

\subsection{Dataset domain}
The data used in this work was collected to study forest dynamics following wildfires. It was collected at four locations in California, USA that experienced major wildfires between two to nine years prior to data collection (\cref{tab:site_attributes}). All sites are dominated by the ``California mixed-conifer'' forest type \cite{barbour2007terrestrial, safford2017natural} containing primarily sugar pine (\emph{Pinus lambertiana}, labeled ``PILA''); white fir (\emph{Abies concolor}, ``ABCO''); Douglas-fir (\emph{Pseudotsuga menziesii}, ``PSME''); incense cedar (\emph{Calocedrus decurrens}, ``CADE''); and ponderosa pine (\emph{Pinus ponderosa}) or Jeffrey pine (\emph{Pinus jeffreyi}) (collectively labeled  ``PIPJ'' due to the inability to definitively distinguish them in the field). The data consists of both drone imagery and manual field reference observations.

\subsection{Field reference data}
The field reference data consists of manual measurements of the geospatial (lat/lon) locations of trees as well as attributes including species and height. In the field, tree locations were in most cases mapped based on their distance (measured by laser rangefinder) and azimuth (measured by compass) relative to survey centerpoints, which were located using a high-precision RTK GPS system. In some areas with sufficient sky view, individual tree locations were instead directly measured with the RTK GPS system. To co-register the field-mapped trees to the drone-derived products (mesh model, orthomosaic, canopy height model, and camera poses), we applied a manually identified x,y shift identically to all trees measured from a given centerpoint (or located directly using RTK GPS) such that prominent field-surveyed trees aligned optimally with prominent drone-detected trees.

\subsection{Drone imagery}

The drone data was collected using a combination of DJI Phantom 4 Advanced and DJI Phantom 4 Pro v2 drones, which both use an integrated 20 megapixel RGB camera (model FC6310/FC6310s) with the same specifications. A flight planner was used to collect images using two different missions per site: one with parameters optimized for structure reconstruction and structure-based tree detection, and one with parameters that were hypothesized to enhance species classification performance. The structure reconstruction mission was executed at 120 m altitude above ground level with a nadir-facing camera and 90\% front and side overlap. The species classification mission was executed as a ``grid'' mission consisting of two perpendicular sub-missions at an altitude of 80 m above ground level (except 90 m for the Valley site) with 80-90\% front and 70-80\% side overlap with a camera pitch of 20 degrees off nadir (except 25 degrees for the Lassic site).

\subsection{Photogrammetry products}
The imagery from each site was processed using Agisoft Metashape version 2.0 or 2.1 \cite{Agisoft2021} using the workflow and parameterization identified by Young \etal \cite{Young2022} as optimal for mapping conifer forests. The published workflow was modified to use an image upscaling factor of 4 (instead of 2) during the depth mapping/dense cloud generation step to account for internal Metashape changes since version 1.6 that was used for the published workflow development. In the meshing step, which is not included in the published workflow, we used the Metashape defaults. Each site processed successfully and all or nearly all of the cameras aligned properly. Manually selected ground control points were used to accurately align the photogrammetry project with geospatial coordinates.

The first output we use is the location of the cameras, which captures the pose of each image as well as the calibrated intriniscs of the camera that are shared across all images. Second, we use the mesh model of the scene in geospatial coordinates. Finally, we use two geospatial raster products: the digital surface model (DSM) represents the elevation of the top surface of the mesh while the digital terrain model (DTM) represents the elevation of the ground surface predicted by Metashape.

\subsection{Tree detection}

Following previous work \cite{Young2022}, treetops were detected as local maxima within the canopy height model (CHM) raster. The CHM was produced by subtracting the DTM from the DSM, resampling the result to 0.25 m resolution via bilinear interpolation, and applying a 7x7 pixel sliding window mean smooth. Trees were detected using a variable-radius sliding window filter on the smoothed CHM, parameterized to use a circular local maximum search window with a horizontal radius equal to 0.11 times the height of the focal pixel. Local maxima with height $>$ 5 m were retained. The local maximum search was performed using the \texttt{locate\_trees} and \texttt{lmf} functions in the R package \texttt{lidR} \cite{lidR}. 

The horizontal extent of the crown surrounding each detected treetop was delineated using the method of Silva \etal \cite{Silva2016Imputation}, parameterized to include all contiguous pixels with a height greater than 0.1 times the treetop height, to a maximum horizontal distance of 0.24 times the treetop height. The method delineates intersecting crowns using a Voronoi tessellation. Crown delineation was performed using the \texttt{segment\_trees} and \texttt{its\_silva2016} functions in the R package \texttt{lidR}. The resulting crown polygons were then simplified using a kernel smoothing filter. Field labels were assigned to detected trees using the algorithm of Young \etal \cite{Young2022} developed for this task (Sup. \cref{suplementary:dataset:matching_trees}).
\section{Multiview Semantic Meshes}
\label{sec:methodology}
\begin{figure*}
  \centering
  \includegraphics[width=\linewidth]{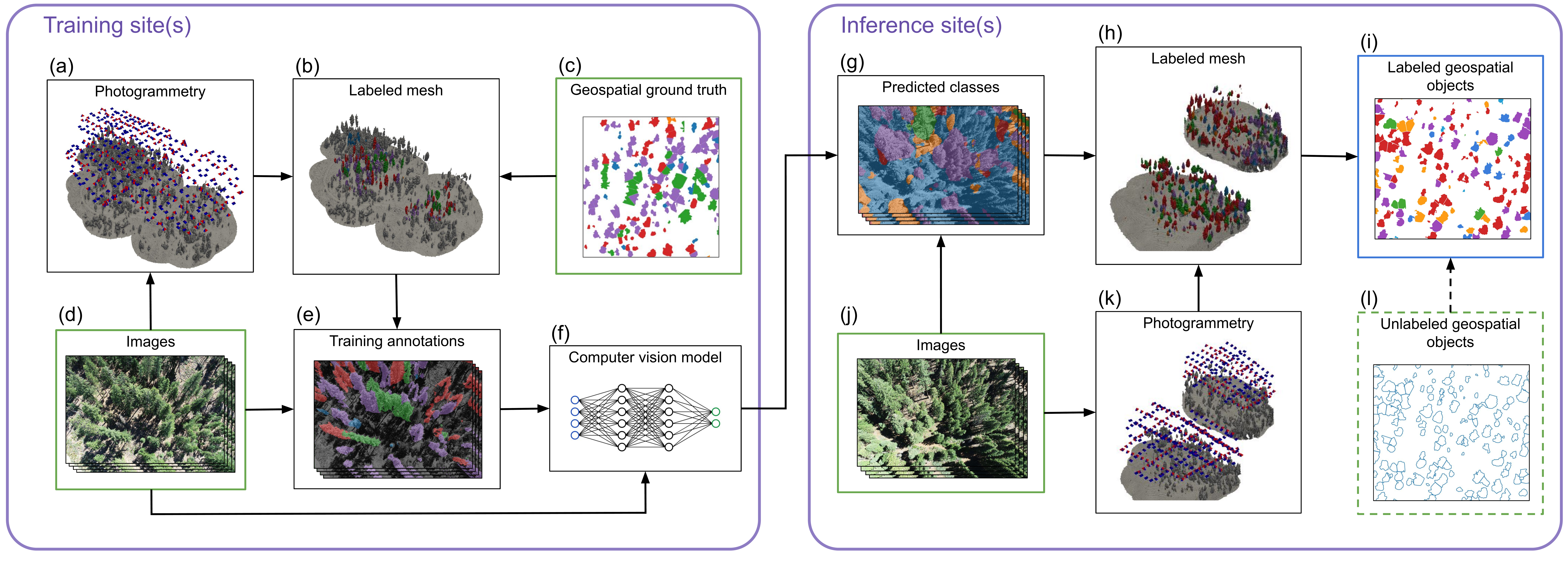}
  \caption{Schematic of the analytical workflow, with input data (c, d, j, l) outlined in green and output data (i) outlined in blue. At the sites used to train the computer vision model, raw drone images (d) are collected and processed using photogrammetry to estimate camera poses and a 3D geospatial mesh model (a), which is then textured using geospatial ground-truth species labels (c) obtained via field surveys. The semantic mesh tool presented here is then used to render species labels (e) that match pixel-for-pixel with the raw drone images. The raw images and labels are then used to train a computer vision semantic segmentation model (f). To classify trees to species at a new site, another set of drone images (j) is collected and segmented pixel-for-pixel into species classes (g) using the trained model. In parallel, the images are processed using photogrammetry to yield camera poses and a 3D mesh model (k). The semantic mesh tool presented here is then used to project the image-based species classes onto the mesh faces (h). The geospatial locations of trees (l) are determined, and the mesh-based species classes are transferred to the geospatial tree map to yield the final output, a tree map with inferred species labels (l).}
  \label{fig:concept}
\end{figure*}

The goal of multiview mapping is to generate predictions for some task on raw images and then convert these predictions into geospatial data that is useful for land managers. In practice, these predictions are generated using deep learning models applied to the images. It is uncommon for pre-trained models to exist for a given ecology task, so the user generally has to train their own domain-specific model. Hand labeling data for model training is laborious, and for many ecology problems it cannot be accurately conducted without relying on in-situ observations. Given these challenges, a significant portion of the multiview mapping workflow is devoted to training models using relevant geospatial ground truth data. 

\subsection{Definitions}
\label{sec:methodology:defs}
\textbf{Class IDs:}
We assume that in our classification problems there are a fixed number of classes $N_{clc}$ that are known beforehand. Therefore, our set of potential $IDs$ is represented by a set of integers, one for each class, plus the special $null$ value used to represent background or unlabeled content. This is represented as $IDs = \{1 ... N_{cls}\} \cup \{null\}$.\\ 
\textbf{Geospatial labels:}  
The ground truth data used to train machine learning models is provided in geospatial format. Specifically, this data consists of a set of closed 2D polygons, each with an associated ID in $IDs$. These polygons are assumed to be non-overlapping. The set of geospatial label polygons is represented as $L_g$, and $L_g[x,y]$ can be used to query the class ID for a given 2D geospatial location. If the point is not contained within any of the polygons, the $null$ class ID is returned. \\
\textbf{Class ID images:}
This approach currently assumes that machine learning models are trained on and predict per-pixel integer labels for each image they receive,  represented as a $I_{h} \times I_{w}$ array of integer class IDs with the same number of rows and columns as the original image. If this label image array is defined as $L_i$, the image can be indexed with an integer pixel location to obtain the class for that pixel as $L_i[i,j]$. Any indices outside the bound of the image return the $null$ label.\\
\textbf{Meshes:}
The 3D geometry of the scene is represented by a mesh consisting of a set of points called \textit{vertices} and a set of triangular \textit{faces}. The vertices $V$ are represented as an $N_{v} \times 3$ array, where each row represents a point in 3D space. The faces $F$ are represented by an $N_{f} \times 3$ array, where each row consists of three non-repeating integer indices into the vertex array describing the corners of a triangular face. \\
\textbf{Cameras:}
Virtual cameras are used to represent the properties of how each image was captured. These cameras are primarily parameterized by $4\times4$ extrinsic matrix, representing the camera's location plus orientation, and a $3\times3$ intrinsic matrix, representing projection of 3D points in the camera's field of view into the 2D location of pixels. Both of these matrices are estimated by photogrammetry software. A thorough review of geometric computer vision concepts is provided by Hartley and Zisserman \cite{Hartley2004}. We represent the set of cameras for a scene as $C$ with $N_{cam}$ elements and define the individual cameras as $c \in C$.\\
\textbf{Digital terrain models:}
The digital terrain model represents the estimated elevation of the ground surface and is represented as $G$. The estimated elevation can be queried for a given 2D location to obtain the elevation of the ground at that point as $G[x,y]$. The ground surface estimate can be produced by many photogrammetry software and in this work is assumed to be defined for the entire study region.

\subsection{Computing image-to-mesh correspondences}
The core operation in the semantic mesh framework is computing which face on the mesh is observed by a given pixel in a camera's image. This involves using the camera's extrinsic and intrinsic matrices and the pixel indices to define a ray originating at the camera's location and extending in the direction of that pixel's observation. Then, the index of first mesh face that the ray intersects is reported. For a mesh $M$, camera $c$, and pixel index $(i,j)$, \cref{eq:pix2face} is used to compute a face index $F_i$ representing which face would be observed from that pixel. If the ray does not intersect any face, which can occur in the case of cropped meshes or inaccurate 3D reconstruction, the $null$ index is returned. 
\begin{align}
    I_f = pix2face(M, c, (i,j)) \label{eq:pix2face}
\end{align} 
There are numerous ways to implement the $pix2face$ function using modern computer graphics libraries. In this work, we use the explicit pixel-to-face correspondences computed in the \texttt{PyTorch3D} \cite{ravi2020pytorch3d} differentiable rendering pipeline.

\subsection{Training workflow}
\label{sec:methodology:training}
The training workflow generates per-image training labels using a mesh $M$, set of cameras $C$, ground plane estimate $G$, and geospatial polygon labels $L_g$.

The region observed in the aerial survey is often significantly larger than the region labeled in the field survey. For computational reasons, we define a 2D region of interest around the labels and only consider this area during the training label generation process. Specifically, we define the $RoI$ as a closed 2D polygon consisting of locations within a user-defined buffer radius $r_{b}$ of any polygon in $L$. The user should set $r_b$ conservatively, such that it is unlikely any cameras outside of this radius can observe labeled regions. The mesh $M$ is updated so that only faces contained entirely within $RoI$ are retained. Similarly, the camera set $C$ is filtered so that only cameras located within the $RoI$ are kept.

The next step is to add class IDs to $M$ using the information contained in $L$. The label for each vertex in the mesh vertices $V$ is obtained using \cref{eq:vert_labels}.
\begin{align}
L_v &= \{L_g[v_x, v_y] \mid v \in V\} \label{eq:vert_labels}
\end{align}

Since the per-vertex labeling is performed using a 2D polygon, some regions of ground below tree canopies may be labeled as the class of the canopy above. To filter out these spurious labels, we use a simple height-above-ground filter. The vertex elevation is compared to the estimated ground elevation at that 2D location to obtain a height above ground $H_g$ for each vertex (\cref{eq:ground_height}). The vertices that have a height below the user-defined threshold height $H_t$ are computed and the corresponding vertex labels are set to the $null$ label (\cref{eq:set_ground_null}).

\begin{align}
H_g &= \{v_z - G[v_x, v_y] \mid v \in V\} \label{eq:ground_height} \\
L_v[H_g < H_t] &= null \label{eq:set_ground_null}
\end{align}

Then each face is labeled based on the most common label for the associated vertices (\cref{eq:face_labels}). The $mode$ function is defined to randomly tie-break in situations where all the vertex labels differ. A face label is set to $null$ only if all vertices are labeled $null$.
\begin{align}
L_f &= \{mode(L_v[f_1], L_v[f_2], L_v[f_3]) \mid f \in F\} \label{eq:face_labels} 
\end{align}

\begin{figure}
  \centering
   \begin{subfigure}{\linewidth}
   \includegraphics[width=\linewidth]{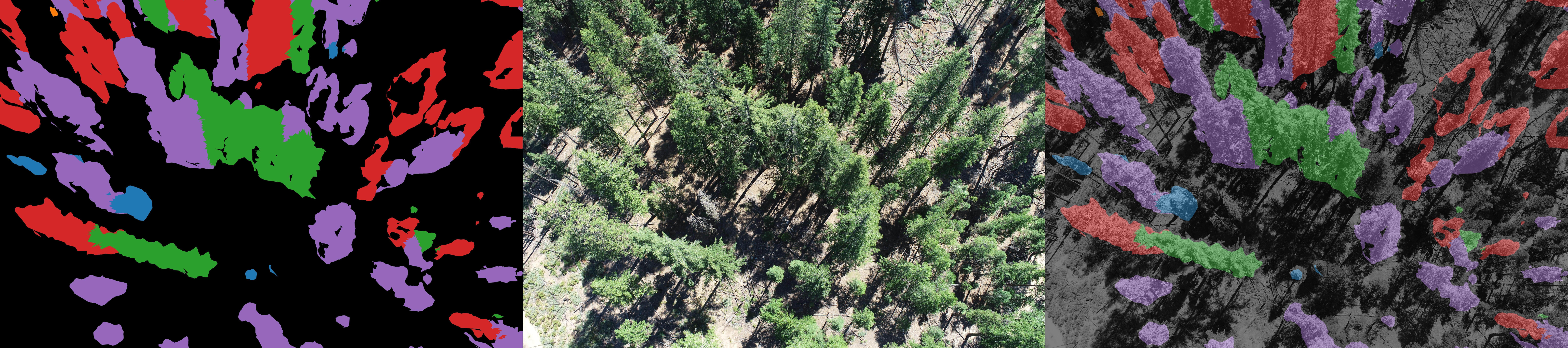}
   \end{subfigure}
   \begin{subfigure}{\linewidth}
   \includegraphics[width=\linewidth]{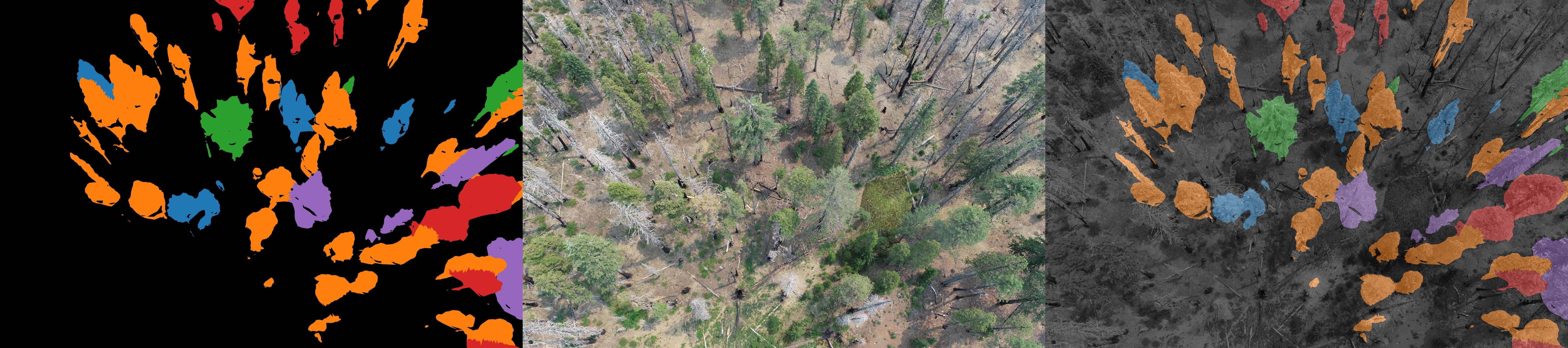}
   \end{subfigure}
  \caption{Labels rendered during the training process. The left pane shows the label IDs colored by class, the middle shows the raw images, and the right shows the classes overlayed on a grayscale image.}
  \label{fig:rendered_labels}
\end{figure}

Using the labeled mesh we generate per-image labels for each virtual camera. For a given camera $c \in C$, we compute a labeled image $L_i$. For each pixel in the image, we compute the face ID $I_f$ using $pix2face$ (\cref{eq:pix2face}). Then we obtain the label by taking the corresponding face label, $L_f[I_f]$. These IDs are assembled into an image of the same dimension as the input image, and can be used as labels for the corresponding real images for model training (Sup. \cref{fig:rendered_labels}).




\subsection{Prediction workflow}
\label{sec:methodology:prediction_workflow}

The prediction workflow takes as input mesh $M$, a ground plane estimate $G$, a set of cameras $C$ and predicted segmentation image $P_{img}$ for each camera $c \in C$. In the case of per-object classification, the boundaries of each unclassified object are defined as a set of polygons, $O_u$. Similar to the training workflow (\cref{sec:methodology:training}), the mesh is cropped and the camera set is filtered to the $RoI$ around the study site.

For each camera $c \in C$, the corresponding predicted label $L_i \in P_I$ is obtained. A $N_f \times N_{cls}$ matrix $P_{face}$ is initialized for each to all zeros to record the predicted class(s) per face. For each camera, for all pixel indices 
$\{(i, j) \mid i \in 1...I_h, j \in 1...I_w \}$, the predicted class at that pixel is obtained (\cref{eq:query_pred}), the corresponding face index is identified (\cref{eq:compute_face}), and the $P_{face}$ is updated (\cref{eq:set_ind}).
\begin{align}
P_{pix} &= P_{img}[i,j] \label{eq:query_pred} \\ 
I_{face} &= pix2face(M, c, (i,j)) \label{eq:compute_face} \\ 
P_{face}[I_{face}, L_{pix}] &= 1 \label{eq:set_ind} 
\end{align}

This procedure is preformed for each camera to obtain $P^i_{face} \mid i \in 1...N_{cam}$ and these matrices are summed element-wise across cameras to obtain $P^{sum}_{face}$. The most common class prediction per face is computed using \cref{eq:most_common_class}.
\begin{align}
P^{ID}_{face} = \underset{col}{\text{argmax}} P^{sum}_{face} \label{eq:most_common_class}
\end{align}

This produces a mesh where the most commonly-predicted class per face is identified.
For tasks where distinct (unclassified) objects are already identified, we generate a class prediction per object. 
For each object, represented by a 2D polygon, we record which mesh faces have an overlapping top-down 2D projection.
Each face is assigned a weight based on the \textit{3D} area of that face in the original mesh. The 3D area, rather than the 2D projected area, is chosen because the near-vertical faces capture useful information from oblique predictions. Faces that are identified as ground are then down-weighted by a user-provided factor.
Finally, using per-face weighting and predicted classes from $P^{ID}_{face}$, the per-class predictions are summed and the object is assigned to the class with the highest value. This process is repeated for each unlabeled object to obtain per-object labels.

\section{Experiments}
\label{sec:experiments}
\subsection{Dataset splits}
The goal of our experiments is to compare an orthomosaic-based approach and a multiview approach on a challenging realistic task. We adopted a cross-site evaluation methodology to avoid overly-optimistic results caused by training and testing on spatially-adjacent data. Specifically, we rendered the labels for each of the sites. Then, we used a leave-one-out strategy and train a model using all but one site, which we then used for testing. This procedure was repeated for each site with both the orthomosaic (``Ortho'') and multiview (``MV'') approaches.

\begin{figure*}
  \centering
   \begin{subfigure}{0.3\linewidth}
   \includegraphics[width=\linewidth]{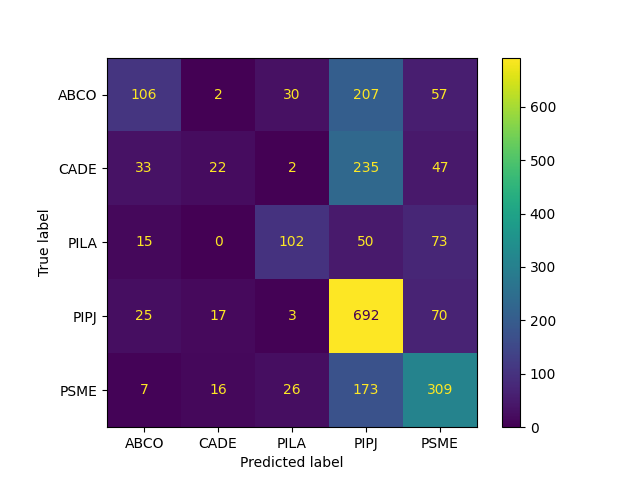}
   \caption{Orthomosaic; accuracy: 53\%.}
   \end{subfigure}
   \begin{subfigure}{0.3\linewidth}
   \includegraphics[width=\linewidth]{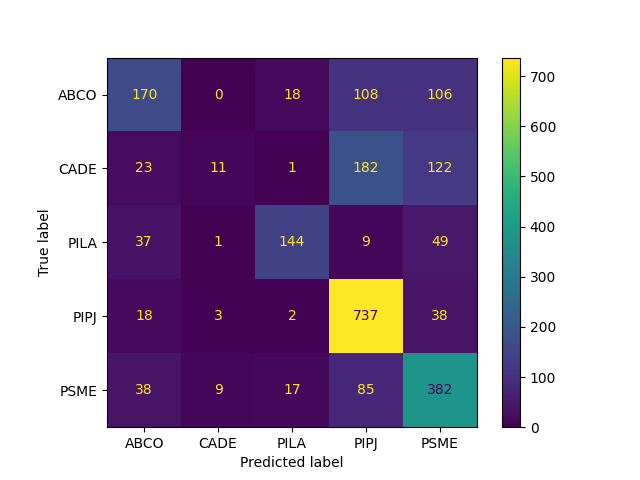}
   \caption{Multiview-HN; accuracy: 63\%.}
   \end{subfigure}
   \begin{subfigure}{0.3\linewidth}
   \includegraphics[width=\linewidth]{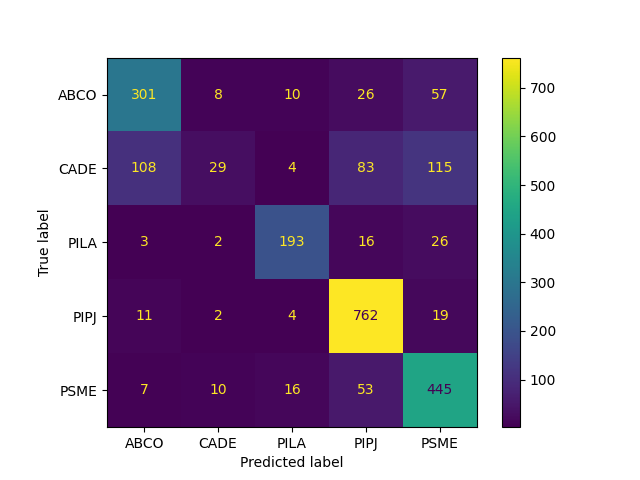}
   \caption{Multiview-LO; accuracy: 75\%.}
   \end{subfigure}
  \caption{Confusion matrices for the cross-site tree species classification task, with species counts summed across the four sites and three trials.}
  
  \label{fig:quantitative_cf}
\end{figure*}
\subsection{Species segmentation with deep learning}
\label{sec:experiments:species_seg}
We treated species classification as semantic segmentation problem where per-pixel labels are produced. To ensure fair comparisons, we used the same deep learning model and training regime in both the multiview and orthomosaic workflows. Specifically, we used the SegFormer \cite{Xie2021} architecture (MIT-B5 backbone variant) implemented in MMSegmentation \cite{mmseg2020} with the default Cityscapes \cite{CordtsCityscapes2016} pretained weights. This architecture was used for a related low-data forest vegetation classification task in the works of Russell \etal \cite{Russell2022Mapping} and Andrada \etal \cite{Andrada2023Semantic} with good results. In all experiments, the models were trained for 10,000 iterations with a fixed learning rate schedule. To address non-determinism in model training, three distinct models were trained on each configuration of data and each used to generated independent predictions, which were then each evaluated. 

\subsection{Orthomosaic baseline}
Our goal was to develop a realistic baseline approach based on an orthomosaic constructed from images taken at 120 m above ground level. Since orthomosaics are often large---frequently 10s of thousands of pixels on a side---tiled model training is critical. In our method we used the orthomosiac at the native resolution and cropped square chips that are the size of the smaller dimension of the images used for photogrammetry and multiview workflow. For training, we used a 50\% overlap between neighboring chips so content at the borders of chips was still used effectively. The classes per pixel were extracted from the corresponding geospatial ground truth polygons and saved as label images with the same width and height as the training crops. For each site, in alphabetical order, 28, 43, 24, and 11, training chips were produced.

We trained models following the procedure outlined in Section \ref{sec:experiments:species_seg}, with three independent models trained on each configuration that excluded a single site. The same procedure to generate orthomosaic chips that was used for training was also used for inference. For each site, the models not trained on any labels from that site were used to generate predictions. These per-chip predictions were then aggregated into a raster prediction for the entire scene. Since the chips overlap, for each pixel in the region, the number of predictions for each class was recorded. To avoid abrupt class transitions at the borders of chips, the pixels in the exterior 25\% were down-weighted using a linear ramp ranging from 1 (full weight) at the interior to 0 (no weight) at the very edge. The most common weighted prediction for each aggregated pixel was reported. Finally, each tree crown polygon was assigned the class that is most commonly predicted within the polygon boundary.
\begin{table*}[h!]
\begin{center}
\begin{tabular}{|| c || c c c | c c c | c c c | c c c | c c c ||} 
 \hline
 & \multicolumn{3}{c|}{Chips} & \multicolumn{3}{c|}{Delta} & \multicolumn{3}{c|}{Lassic} & \multicolumn{3}{c|}{Valley} & \multicolumn{3}{c||}{Aggregated} \\ [0.5ex] 
 & Or. & \makecell{MV-\\HN} & \makecell{MV-\\LO} & Or. & \makecell{MV-\\HN} & \makecell{MV-\\LO} & Or. & \makecell{MV-\\HN} & \makecell{MV-\\LO} & Or. & \makecell{MV-\\HN} & \makecell{MV-\\LO} & Or. & \makecell{MV-\\HN} & \makecell{MV-\\LO} 
 \\ [0.5ex] 
 \hline\hline

Acc. & 0.54 & 0.71 & \textbf{0.82} & 0.60 & 0.71 & \textbf{0.77} & 0.29 & 0.28 & \textbf{0.52} & 0.86 & 0.92 & \textbf{1.00} & 0.53 & 0.63 & \textbf{0.75}\\
std & 0.01 & 0.02 & 0.01 & 0.01 & 0.01 & 0.01 & 0.00 & 0.05 & 0.01 & 0.04 & 0.02 & 0.00 & 0.00 & 0.01 & 0.01\\ \hline
Rec. & 0.41 & 0.53 & \textbf{0.73} & 0.44 & 0.56 & \textbf{0.65} & 0.56 & 0.57 & \textbf{0.73} & 0.92 & 0.89 & \textbf{1.00} & 0.44 & 0.54 & \textbf{0.69}\\
std & 0.05 & 0.03 & 0.00 & 0.01 & 0.01 & 0.01 & 0.02 & 0.03 & 0.04 & 0.05 & 0.01 & 0.00 & 0.00 & 0.01 & 0.01\\ \hline
Prec. & 0.56 & 0.61 & \textbf{0.77} & 0.50 & 0.72 & \textbf{0.73} & 0.47 & 0.50 & \textbf{0.59} & 0.81 & 0.87 & \textbf{1.00} & 0.54 & 0.63 & \textbf{0.72}\\
std & 0.05 & 0.02 & 0.02 & 0.10 & 0.05 & 0.01 & 0.03 & 0.04 & 0.04 & 0.02 & 0.03 & 0.00 & 0.02 & 0.04 & 0.01\\ \hline
 
\end{tabular}
\end{center}
\caption{Leave-one-site-out tree species inference performance. Recall (Rec.) and precision (Prec.) are macro-averaged across species. The aggregated results across the sites are computed by summing the confusion matrices across all sites and then computing the summary statistics. Or.: orthomosaic; MV-HN: High-altitude nadir multiview; MV-LO: Low-altitude oblique multiview. Best values are bold.}
\label{table:per_site_accuracy}
\end{table*}

\subsection{Multiview workflow}
The multiview (``MV'') workflow was performed twice for each site: once using the high altitude (120 m) nadir images (``MV-HN''; the same images used to construct the orthomosaic and mesh), and once using the low altitude (80-90 m) oblique images (``MV-LO''). For each of these two runs of the workflow, the first step was to generate training labels (pixel-wise class annotations) for the raw images from each of the four sites. Since the ground truth polygon data was guaranteed to be non-overlapping, it could be directly used without any pre-processing. The region of interest was defined as any location within 50 meters of the ground truth data, to account for unlabeled trees that could potentially occlude the labeled trees. Following the procedure described in Section \cref{sec:methodology:training}, the mesh vertices and then faces were textured from the ground truth polygon data. Since faces are labeled if they fall within the 2D, top-down boundary of the polygons, some mesh faces corresponding to the ground were labeled with the class of the tree crown above them. To avoid spurious training labels, these faces were filtered out using a simple height-based filter. We used a digital terrain model (DTM) computed during the photogrammetry step to estimate the ground surface and and check the height of each face above the DTM. Faces less than 2 meters above the ground were set to the $null$ class. Finally, the label images were rendered from the perspective of every camera within 50 horizontal meters of the training polygons. The sites, in alphabetical order, had 360, 453, 529, and 125 training images each for the HN dataset and 437, 687, 495, and 143 for the LO dataset. Models were trained using eight separate dataset configurations (four for MV-HN and four for MV-LO), in each case using data from all but one site within the given dataset (MV-HN or MV-LO). For each configuration, three independent models were trained.

To obtain tree class predictions, semantic segmentation was performed for the images at each site using the model trained on the other three sites. Then, these predictions were aggregated onto the mesh using the procedure described in Section \ref{sec:methodology:prediction_workflow}. Ground faces were identified using the same procedure used in training and ground face predictions were down-weighted by a factor of 0.01.

\subsection{Accuracy assessment methods}
For each of the four sites, one set of predictions was generated using the orthomosaic and two using the multiview approach (one for MV-HN and one for MV-LO). For each configuration, we generated three independent predictions using each of the replicated models. For each set, we computed the binary recall and binary precision per class and macro-averaged these metrics across classes to highlight the performance on rarer classes. We also aggregated (summed) the confusion matrices across all sites and all replications and using these confusion matrices computed the overall accuracy, recall (macro-averaged by class) and precision (also macro-averaged by class). In all cases, the site that was predicted was excluded from the training data, and the orthomosaic, MV-HN, and MV-LO datasets were always kept separate.

\subsection{Results}
The low-oblique multiview (MV-LO) approach performed better than the orthomosaic approach in every instance of comparison (\cref{table:per_site_accuracy}). On the aggregate metrics, the MV-LO approach conferred a major improvement in accuracy (75\% vs. 53\%), recall (69\% vs. 44\%), and precision (72\% vs. 54\%) relative to the orthomosaic approach. The high-nadir multiview (MV-HN) approach performed worse than the MV-LO approach (aggregate accuracy: 63\% vs. 75\%; \cref{table:per_site_accuracy}), as expected given that higher-altitude images capture less fine detail and and that nadir images do not capture clear side views that can aid species differentiation. Nonetheless, even the MV-HN approach generally performed substantially better than the orthomosaic approach (aggregate accuracy: 63\% vs. 53\%). The MV-HN approach relied on the same images that were used to create the orthomosaic, so its improved performance demonstrates that classification tasks can benefit from (a) using multiple views, such as the slightly oblique perspectives that are present at the edges even of nadir-facing images, (b) using raw images that contain less distortion than the orthomosaic, or (c) both.

\begin{figure*}
  \centering
   \begin{subfigure}{0.22\linewidth}
   \includegraphics[width=\linewidth]{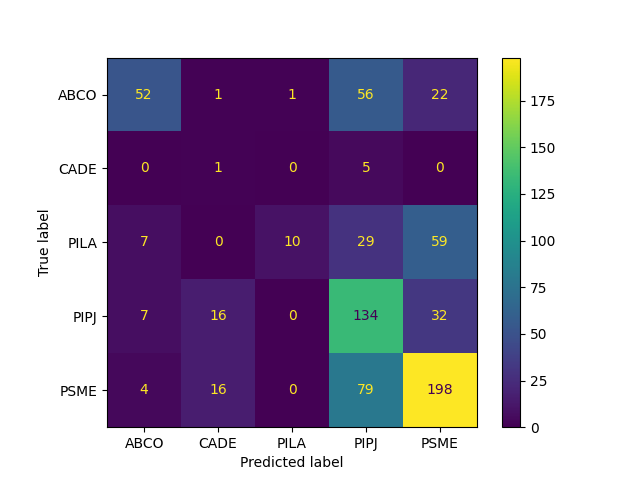}
   \caption{Chips ortho. Acc: 54\%}
   \end{subfigure}
   \begin{subfigure}{0.22\linewidth}
   \includegraphics[width=\linewidth]{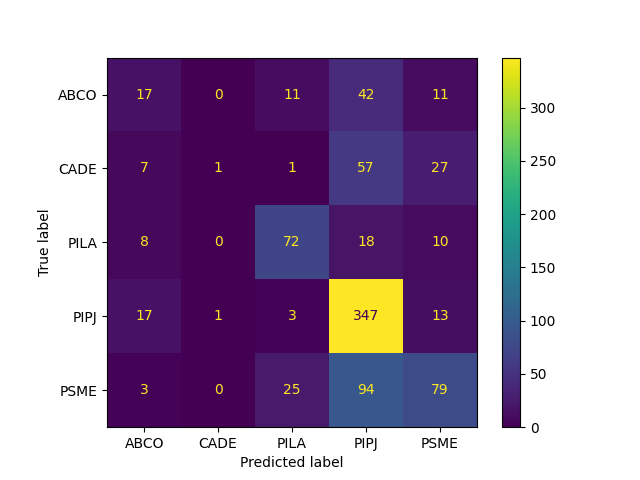}
   \caption{Delta ortho. Acc: 60\%}
   \end{subfigure}
   \begin{subfigure}{0.22\linewidth}
   \includegraphics[width=\linewidth]{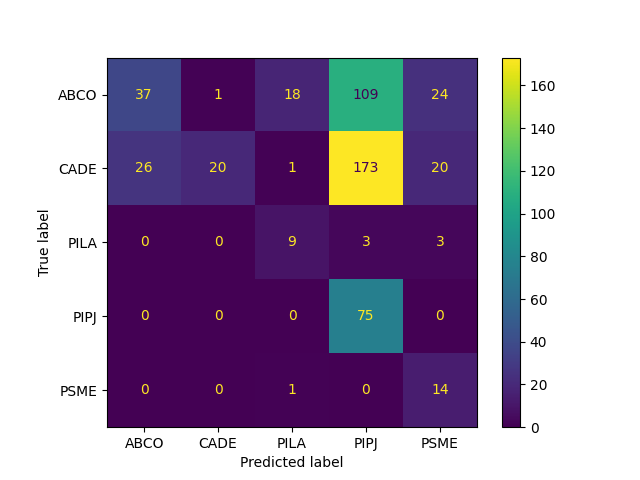}
   \caption{Lassic ortho. Acc: 29\%}
   \end{subfigure}
   \begin{subfigure}{0.22\linewidth}
   \includegraphics[width=\linewidth]{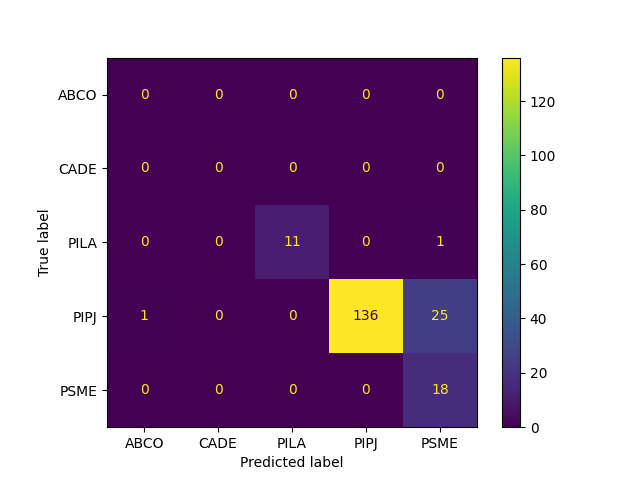}
   \caption{Valley ortho. Acc: 85\%}
   \end{subfigure}

   \begin{subfigure}{0.22\linewidth}
   \includegraphics[width=\linewidth]{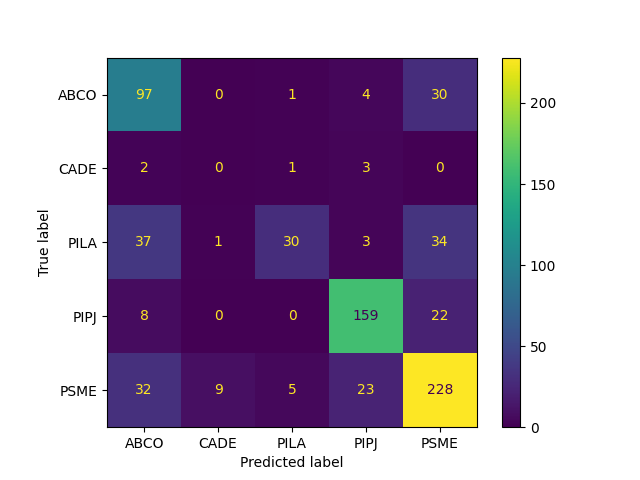}
   \caption{Chips MV-HN. Acc: 70\%}
   \end{subfigure}
   \begin{subfigure}{0.22\linewidth}
   \includegraphics[width=\linewidth]{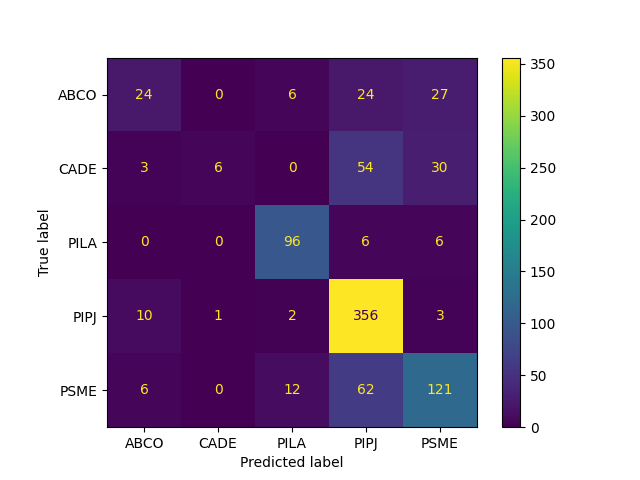}
   \caption{Delta MV-HN. Acc: 70\%}
   \end{subfigure}
   \begin{subfigure}{0.22\linewidth}
   \includegraphics[width=\linewidth]{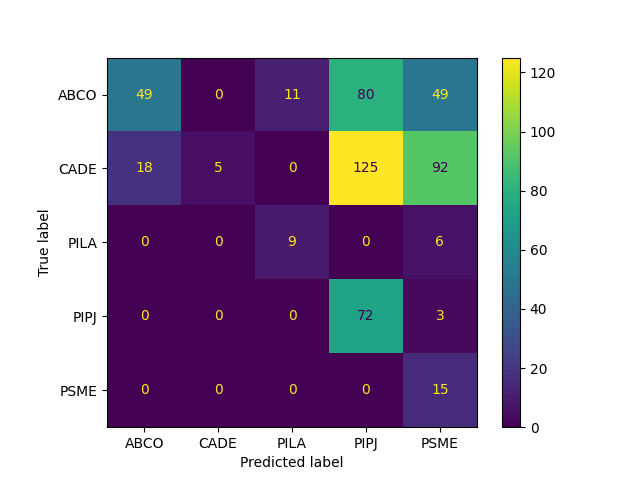}
   \caption{Lassic MV-HN. Acc: 28\%}
   \end{subfigure}
   \begin{subfigure}{0.22\linewidth}
   \includegraphics[width=\linewidth]{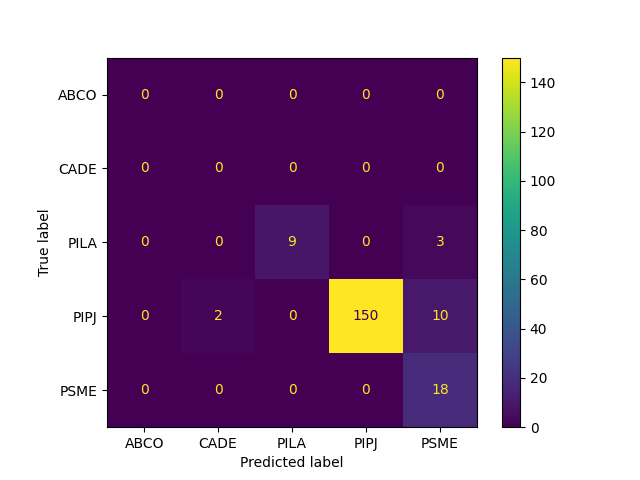}
   \caption{Valley MV-HN. Acc: 92\%}
   \end{subfigure}

   \begin{subfigure}{0.22\linewidth}
   \includegraphics[width=\linewidth]{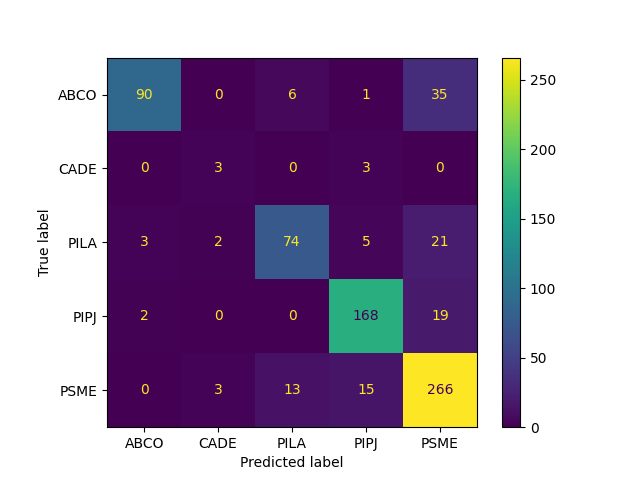}
   \caption{Chips MV-LO. Acc: 82\%}
   \end{subfigure}
   \begin{subfigure}{0.22\linewidth}
   \includegraphics[width=\linewidth]{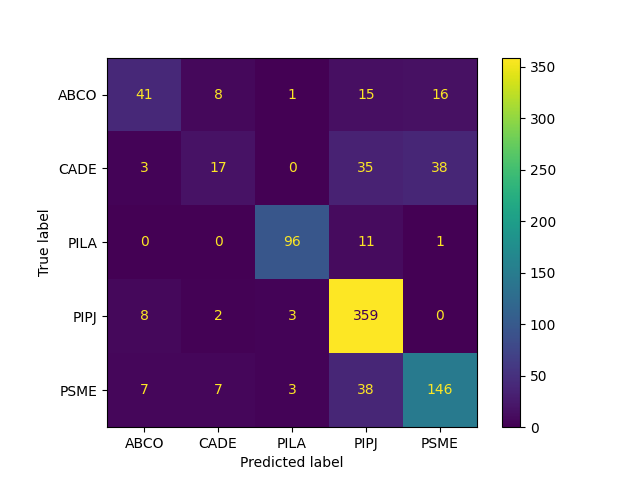}
   \caption{Delta MV-LO. Acc: 77\%}
   \end{subfigure}
   \begin{subfigure}{0.22\linewidth}
   \includegraphics[width=\linewidth]{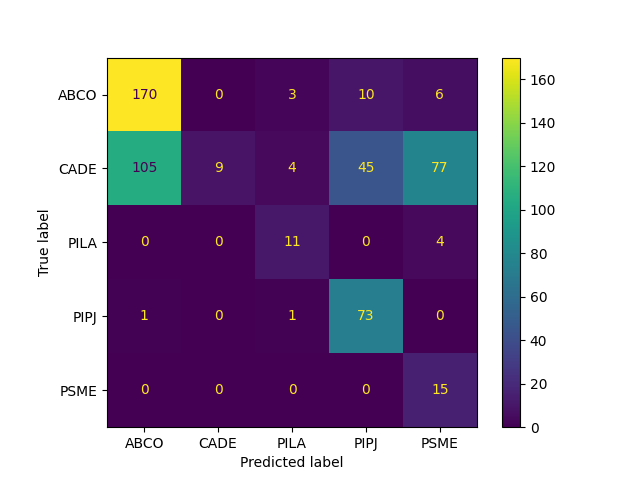}
   \caption{Lassic MV-LO. Acc: 52\%}
   \end{subfigure}
   \begin{subfigure}{0.22\linewidth}
   \includegraphics[width=\linewidth]{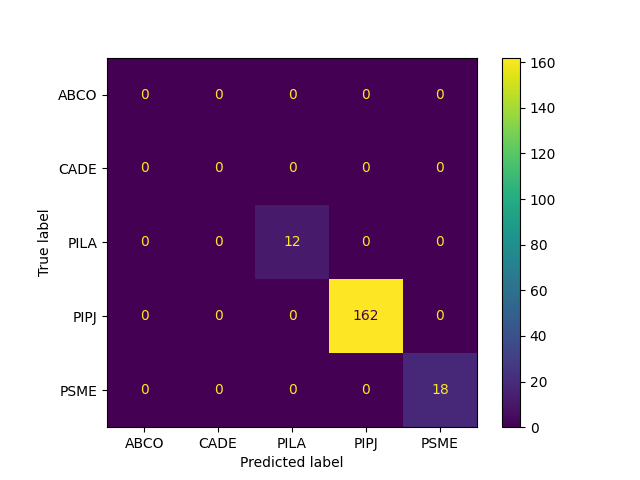}
   \caption{Valley MV-LO. Acc: 100\%}
   \end{subfigure}

  \caption{Site-level confusion matrices for the leave-one-site-out tree species classification task, summed over all trials, for the orthomosaic (ortho.) dataset (a-d), high-nadir multiview (MV-HN) dataset (e-h), and low-oblique multiview (MV-LO) dataset (i-l).}
  
  \label{fig:all_cf_matrices}
\end{figure*}

The orthomosaic approach dramatically over-predicted the two most common classes, PIPJ and PSME, while the distribution predicted by the multiview approaches matched the true distribution much more closely (\cref{fig:quantitative_cf}). This pattern generally held across individual sites (\cref{fig:all_cf_matrices}). Species-level recall and precision varied substantially depending on the species---likely due to variation in training sample size and distinctiveness of appearance---but were consistently higher for the multiview approaches (\cref{table:class_accuracy}). With the MV-LO approach, recall for the most common class (PIPJ) was very high (95\%), and precision was also reasonably high (81\%) and higher than for the less common classes. This was less true for the orthomosaic approach, suggesting that the multiview approach helps to constrain the overprediction of the most common classes.
\section{Conclusions}
\label{sec:conclusions}
This study demonstrates the value of multiview reasoning and deep learning applied to raw drone images for geospatial prediction tasks. The improvement in tree species classification performance that multiview reasoning yielded in our experiments may result from a combination of several factors, including elimination of artifacts from the orthomosaicing process, increased number and diversity of training examples, improved robustness due to prediction ensembling, and added information from the side views of objects. Further research may help to understand the contributions of each of these mechanisms.

Although forest stand structure may be best reconstructed from high-altitude (e.g., 120 m above ground level) drone imagery \cite{Young2022}, our work suggests that to obtain optimal species classification results, an additional imaging mission to collect lower-altitude, oblique imagery should be performed. However, when performing a second mission is infeasible, the multiview approach applied to the same high-altitude nadir images used for structure estimation can still enable increased accuracy relative to an orthomosaic-based approach.

The workflow we present is applicable to land cover mapping tasks and to classification of existing polygons, but it cannot be naively applied to object detection tasks. Future work will extend this framework to multiview object detection. The large, novel forest dataset underlying our experiments is available to support such future development. We aim for it, along with our open-source toolbox, to enable researchers and practitioners in forestry and other geospatial domains to leverage and continue to develop multiview reasoning tools.
{
    \small
    \bibliographystyle{ieeenat_fullname}
    \bibliography{main}
}

\clearpage
\setcounter{page}{1}
\maketitlesupplementary
\section{Dataset}
\label{supplementary:dataset}
\subsection{Photogrammetry}
\label{supplementary:dataset:photogrammetry}
\begin{itemize}
    \item Metashape version: 2.0, except 2.1 for the Lassic site and for aligning oblique images to the nadir project. Platform: Ubuntu 22.04, AMD EPYC-Milan CPU and NVIDIA GRID A100X-40C GPU via IU Jetstream2 cloud compute instances
    \item GCP details: To ensure proper geospatial registration of the photogrammetry products, we used a post-survey ground control point (GCP) workflow. We located distinctive point features (e.g., the corner of a boulder or intersection of two logs) in Google Earth imagery (at least 5 per site) and recorded their geospatial coordinates, as well as the elevation from a USGS 3DEP 10 m DEM at each location. We then located those same landscape features in at least 3 drone images each and recorded their pixel coordinates. This process was performed for the 120 m image set for each site. Tabular GCP data are provided within each site's 120 m imagery mission folder, in the folder named `gcps'. These data are provided in the format expected by the \texttt{automate-metashape}\footnote{\url{https://github.com/open-forest-observatory/automate-metashape}} library v0.2.0. 
\end{itemize}
\subsection{Matching detected and field reference trees}
\label{suplementary:dataset:matching_trees}

Tree species labels recorded during the field surveys were assigned to the tree crowns detected from the drone-derived data using the algorithm of Young \etal \cite{Young2022} developed for this task. The pairwise horizontal distances between all field-mapped tree stems and drone-mapped treetops were computed, and candidate field-drone pairings were retained if (a) the height of the drone tree was within $\pm 50\%$ of the height of the field tree, and (b) the horizontal distance was less than $0.1h + 1$, where $h$ is the height of the field tree and units are in meters. Among the remaining candidate pairings, final pairs were determined in order of increasing distance, with the constraint that once a tree is assigned to a final pairing, it cannot be included in any other pairings.


\section{Supplemental figures and tables}
\label{supplementary:additional_analysis}
Included below.
\begin{table*}[h!]
\begin{center}
\begin{tabular}{||c || ccc | ccc | ccc | ccc || ccc ||} 
 \hline
 & \multicolumn{3}{c|}{ABCO (128)} & \multicolumn{3}{c|}{CADE (104)} & \multicolumn{3}{c|}{PILA (78)} & \multicolumn{3}{c||}{PIPJ (250)} & \multicolumn{3}{c|}{PSME (176)} \\ [0.5ex] 
 & Or. & \makecell{MV-\\HN} & \makecell{MV-\\LO} & Or. & \makecell{MV-\\HN} & \makecell{MV-\\LO} & Or. & \makecell{MV-\\HN} & \makecell{MV-\\LO} & Or. & \makecell{MV-\\HN} & \makecell{MV-\\LO} & Or. & \makecell{MV-\\HN} & \makecell{MV-\\LO} \\ [0.5ex] 
 \hline\hline
Rec. & 0.26 & 0.42 & \textbf{0.75} & 0.06 & 0.03 & \textbf{0.09} & 0.42 & 0.60 & \textbf{0.80} & 0.86 & 0.92 & \textbf{0.95} & 0.58 & 0.72 & \textbf{0.84} \\
std & 0.04 & 0.09 & 0.02 & 0.01 & 0.00 & 0.02 & 0.02 & 0.04 & 0.02 & 0.00 & 0.01 & 0.01 & 0.04 & 0.02 & 0.01\\ \hline
Prec. & 0.57 & 0.61 & \textbf{0.70} & 0.41 & 0.51 & \textbf{0.57} & 0.63 & 0.80 & \textbf{0.85} & 0.51 & 0.66 & \textbf{0.81} & 0.56 & 0.55 & \textbf{0.67}\\
std & 0.02 & 0.06 & 0.02 & 0.13 & 0.21 & 0.03 & 0.05 & 0.05 & 0.05 & 0.01 & 0.04 & 0.01 & 0.01 & 0.01 & 0.02\\ \hline

\end{tabular}
\end{center}
\caption{Class-level classification performance in the leave-one-site-out task, with values macro-averaged across sites and standard deviations taken across runs. The number of trees per class is shown in parentheses. Species classes are defined in \cref{sec:dataset}.}
\label{table:class_accuracy}
\end{table*}

\end{document}